\documentclass[runningheads]{llncs}
\usepackage[T1]{fontenc}
\usepackage{graphicx}
\usepackage{amsmath}
\usepackage{booktabs}
\usepackage{multirow}
\usepackage{caption}
\usepackage{float}
\usepackage{placeins}
\usepackage{color, colortbl}
\usepackage{stfloats}
\usepackage{enumitem}
\usepackage{tabularx}
\usepackage{xstring}
\usepackage{xspace}
\usepackage{url}
\usepackage{subcaption}
\usepackage{xcolor}
\usepackage{CJK}
\usepackage{verbatim}
\usepackage{amsmath}
\usepackage{amssymb}
\usepackage{bm}
\usepackage{booktabs} 
\usepackage{multirow}
\usepackage{graphicx}
\usepackage{wrapfig}
\usepackage{array}   
\usepackage{wrapfig}
\usepackage{ulem}
\usepackage{CJKulem}
\usepackage{soul}
\usepackage{hyperref}
\title{On the Interplay of Human-AI Alignment, Fairness, and Performance Trade-offs in Medical Imaging}
\titlerunning{Aligner}

\begin{document}
%

\author{%
  Haozhe Luo\inst{1,3}
  \and Ziyu Zhou\inst{2}
  \and Zixin Shu\inst{1}
  \and Aurélie Pahud de Mortanges\inst{1}
  \and Robert Berke\inst{3}
  \and Mauricio Reyes\inst{1}
}
\authorrunning{H.\ Luo et al.}
\institute{%
ARTORG Center for Biomedical Engineering Research, University of Bern\\
  \and
 Shanghai Jiao Tong University\\
  \and
 kaiko.ai \\
}

\maketitle              

\begin{abstract}

Deep neural networks excel in medical imaging but remain prone to biases, leading to fairness gaps across demographic groups. We provide the first systematic exploration of Human-AI alignment and fairness in this domain. Our results show that incorporating human insights consistently reduces fairness gaps and enhances out-of-domain generalization, though excessive alignment can introduce performance trade-offs,  emphasizing the need for calibrated strategies.  These findings highlight Human-AI alignment as a promising approach for developing fair, robust, and generalizable medical AI systems, striking a balance between expert guidance and automated efficiency. Our code is available at https://github.com/Roypic/Aligner.

\keywords{Fairness  \and Human-AI Alignment  \and Vision Language Model.}
\end{abstract}
\section{Introduction}

Deep neural networks have become indispensable for a wide range of medical image computing applications. Nevertheless, their data-driven nature renders them susceptible to learning spurious correlations and biases \cite{geirhos2020shortcut,degrave2021ai,gichoya2022ai}. This susceptibility not only undermines robustness and generalization, especially under out-of-distribution (OOD) conditions, but also raises concerns about the fairness of these systems across diverse patient populations. Human-AI alignment has recently emerged as a promising avenue for mitigating such issues by directing the learned representations toward human-centric knowledge. Although existing work on Human-AI alignment - also referred to as Explanation-Guided Learning (EGL) \cite{gao2024going,gao2022res,gao2022aligning,rieger2020interpretations,wang2022follow,wu2024gaze,rieger2020interpretations} - has shown improved robustness and performance, its relationship with model fairness remains largely unexplored.
In medical imaging, unfairness often manifests as systematic performance disparities across demographic subgroups (e.g., sex, race, age), stemming from biases in training data and inconsistencies in annotation practices, among other factors. For instance, several studies have revealed fairness gaps in chest X-ray classifiers \cite{seyyed2020chexclusion,seyyed2021underdiagnosis}, racial disparities in brain image analysis \cite{stanley2022fairness,jones2024causal}, and gender imbalances yielding skewed diagnostic outcomes \cite{larrazabal2020gender}. Other lines of research highlight unfairness resulting from socioeconomic biases \cite{obermeyer2019dissecting} or presentation and annotation disparities \cite{glocker2021algorithmic,zhang2024data}. Given the potential of Human-AI alignment to mitigate these issues, its impact on reducing fairness gaps warrants deeper investigation.
In this paper, we investigate the interplay between Human-AI alignment and model fairness, a relationship that remains largely unexplored. Specifically, we ask: \textit{“Does Human-AI alignment contribute to reducing disparities in trained models?”}. To this end, we design a study on disease classification from chest X-ray images, a commonly benchmarked task for fairness research for which associated fairness variables are available. We systematically analyze fairness with respect to two subgroups (gender and age), using multiple group fairness metrics. Our experiments are conducted on Vision Transformer (ViT)  under various degrees of human-AI alignment (including deliberate misalignment). 
Our findings demonstrate that Human-AI alignment consistently reduces fairness gaps across diseases and demographic subgroups while also enhancing out-of-domain generalization. This supports recent studies \cite{stanley2022fairness,jones2024causal,zhang2024data} suggesting that mitigating spurious correlations can improve real-world performance, challenging the notion that fairness interventions necessarily degrade model accuracy. However, we also find that excessive or misguided alignment can introduce trade-offs, emphasizing the need for carefully calibrated strategies. To the best of our knowledge, this is the first systematic study of Human-AI alignment’s impact on fairness in medical imaging, highlighting its potential to develop fair, robust, and generalizable AI models.

\section{Methods}
\subsection{Experimental Design to Assess Impact of Human-AI and Fairness}
\begin{figure}[htb]
    \setlength{\belowcaptionskip}{0.1cm}
    \centering
    \includegraphics[width=13cm]{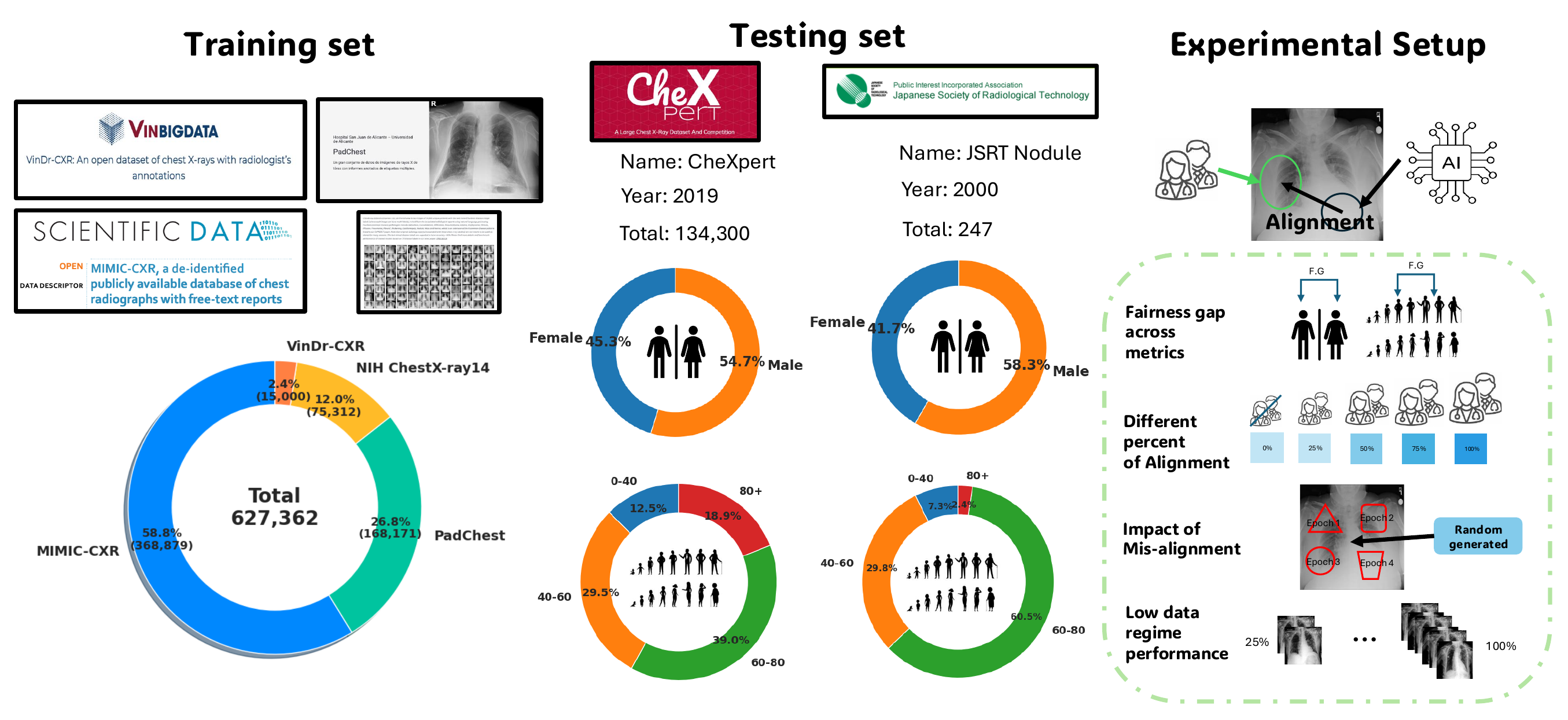}
    \caption{Experimental setup to assess the impact of Human-AI alignment on fairness. Models trained on multicenter training datasets are trained without Human-AI alignment (i.e., baseline), and with various degrees of Human-AI alignment, and their fairness gap and classification performance metrics are assessed across two demographic groups on out-of-domain datasets across three different classification tasks. Additionally, the impact of Human-AI attention on low-data regimes and when alignment is randomized are further evaluated.} 
    \label{fig:studydesign} 
\end{figure}

Figure \ref{fig:studydesign} summarizes our study design, describing the multi-center training and out-of-domain (OOD) data used in our experiments, comprising fairness attributes (sex and age), different levels of human-AI alignment (including a randomized alignment ablation), and evaluation metrics for fairness, and performance (including a subanalysis at different regimes of training data).

\textbf{Training and OOD Data:} We selected multiple publicly available chest X-ray datasets to train classification models for detecting (i) nodules and masses, (ii) pleural effusion, and (iii) edema. These conditions were chosen for their clinical relevance and their distinct semantic characteristics, e.g., spatial location is a key factor for nodules, whereas texture plays a crucial role in identifying pleural effusion. These training datasets come equipped with expert-based annotations reflecting human-based attention areas a radiologist uses for diagnosis. These areas were used to guide the learning process, as detailed in section \ref{sec:HAI}.
Table \ref{tab:TrainOODDataset} provides details on the datasets used for training and OOD evaluation, including NIH ChestX-Ray14 \cite{wang2017chestx}, MIMIC-CXR \cite{johnson2019mimic}, VinDr-CXR~\cite{nguyen2022vindr}, PadChest~\cite{bustos2020padchest}, CheXpert~\cite{irvin2019chexpert}, and CheXlocalize~\cite{saporta2022benchmarking}. These datasets ensure a diverse and comprehensive representation of different pathologies, with a total of $627,362$ cases used for training and $134,547$ for OOD testing.



\begin{table}[ht]
    \centering
    \caption{Datasets used for training and out-of-distribution (OOD) testing. The CheXlocalize dataset is utilized as an additional attention evaluation set of the CheXpert dataset.}
    \label{tab:TrainOODDataset}
    \begin{tabular}{lcccc|cc}
        \toprule
        \textbf{Condition} & \multicolumn{4}{c|}{\textbf{Training }} & \multicolumn{2}{c}{\textbf{OOD Testing}} \\
        \cmidrule(lr){2-5} \cmidrule(lr){6-7}
        & \textbf{NIH} & \textbf{PadChest} & \textbf{VinDr} & \textbf{MIMIC} & \textbf{JSRT} & \textbf{CheXpert} \\
        & \textbf{ChestX-ray14} &   & \textbf{CXR} & \textbf{CXR} &  & \textbf{(CheXlocalize)} \\
        \midrule
        Nodule \& Mass & \checkmark & \checkmark & \checkmark & -- & \checkmark & -- \\
        Pleural Effusion & \checkmark & -- & \checkmark & \checkmark & -- & \checkmark \\
        Edema & \checkmark & -- & \checkmark & \checkmark & -- & \checkmark \\
        \bottomrule
    \end{tabular}
\end{table}

\textbf{Varying degrees of Human-AI alignment:} We investigated the impact of Human-AI alignment on fairness by systematically varying the degree of Human guidance. Specifically, we considered five levels: 
\textit{Level-1}: \textbf{0\% (No Alignment):} A fully data-driven approach where \textbf{no Human-AI alignment} is conducted. \textit{Level-2}: \textbf{25\% (Weak Alignment):} Human-labeled data constitutes 25\% of the total dataset used for standard training. \textit{Level-3}: \textbf{50\% (Moderate Alignment):} Human guidance is incorporated at a level equal to 50\% of the dataset used for standard training. \textit{Level-4}: \textbf{75\% (Strong Alignment):} A predominantly Human-aligned setting where 75\% of the dataset used for standard training consists of Human-labeled data. \textit{Level-5}: \textbf{100\% (Full Alignment):} A model trained on the same total dataset size as the standard cross-entropy baseline, but with complete Human guidance.

\textbf{Varying training dataset size regimes:} We also performed a comparative analysis of fairness between fully Human-AI aligned models and non-aligned models across different training data ratios (i.e., 25\%, 50\%, 75\% and 100\%), where we measured group performance disparities across four key metrics: Accuracy, AUC, F1-score, and True Positive Rate (Sensitivity).

\textbf{Ablation study - Effect of Random Alignment:} We also performed an experiment where we randomized the attention areas the models are promoted to be aligned to. For each epoch, we randomly generate different shapes of attention maps at random locations.

\textbf{Evaluation metrics:}
On the OOD datasets, we assessed fairness using the fairness gap metric proposed in \cite{ktena2024generative}, which quantifies the disparity in AUC performance between the best- and worst-performing demographic subgroups (e.g., male vs. female, and across different age subgroups; see Fig.~\ref{fig:studydesign} for details on subgroup definitions). Following \cite{ktena2024generative}, we considered AUC performance disparity as most relevant given that the positive and negative ratio of samples across all conditions is imbalanced. 
In addition, we evaluated the performance of models using the F1 score, accuracy, area under the ROC curve (AUC), and sensitivity across classes. Finally, to assess the degree of Human-AI alignment, we assessed the level of hit rate, as proposed in the XAI literature \cite{saporta2022benchmarking}.


\subsection{Human-AI Alignment for ViTs}\label{sec:HAI}

\begin{figure}[tb]
    \setlength{\belowcaptionskip}{0.1cm}
    \centering
    \includegraphics[width=12.5cm]{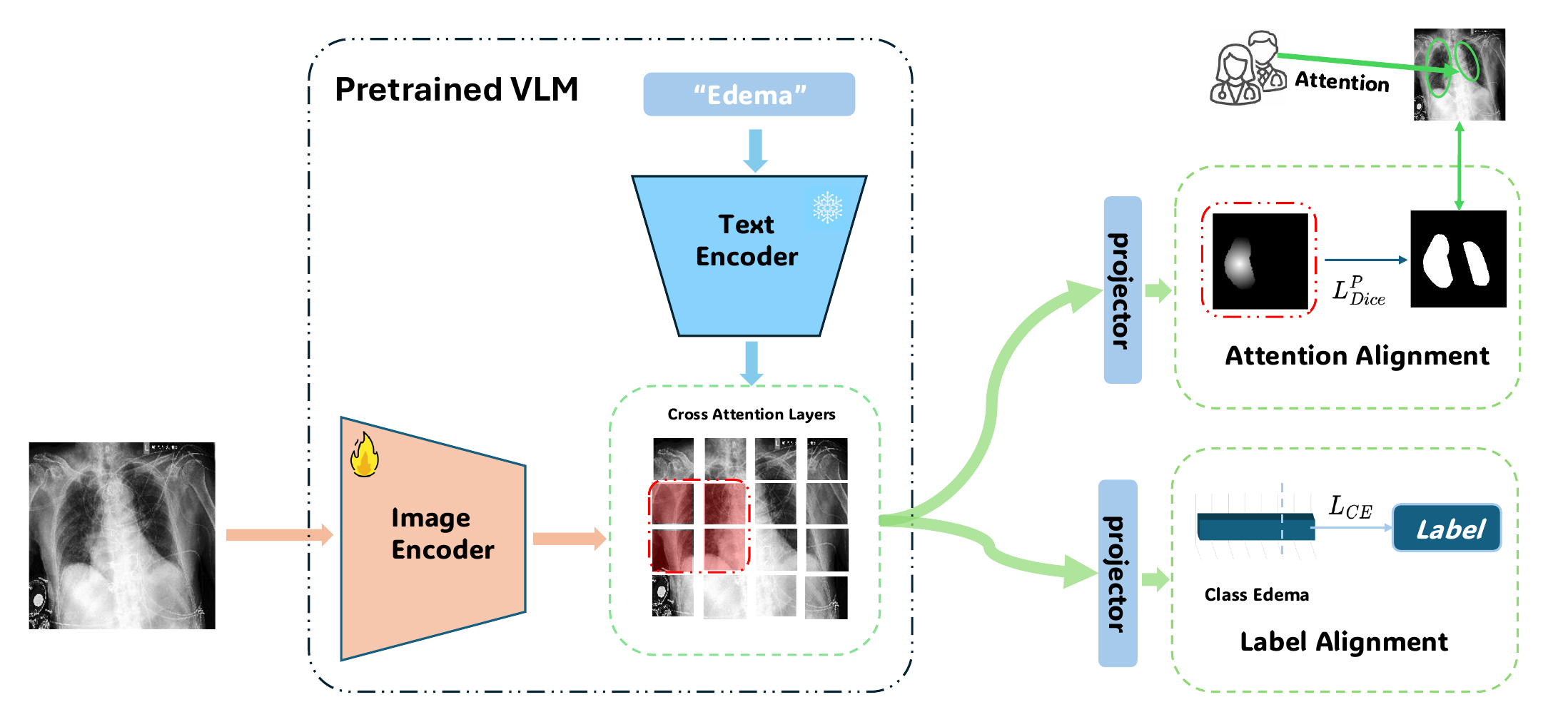}
    \caption{Human-AI Alignment flow chart adapted from \cite{luo2024dwarf}. The approach is based on Visual-Language-Model (VLM) fusing image and language embeddings via cross-attention. The model is trained sequentially for each class per epoch, with the disease name as a prompt (e.g., "Edema"). Two projector heads are used to (i) optimize Human-AI alignment, and (ii) perform disease classification.}
    \label{fig:architecture}
\end{figure}

Fig.~\ref{fig:architecture} illustrates the architecture employed to perform Human-AI Alignment. It builds upon a recently proposed pre-trained medical Vision-Language Model (VLM) for chest X-ray diagnosis \cite{luo2024dwarf}. Below, we provide a summary of the approach for completeness reasons and derive the reader to \cite{luo2024dwarf} for details.

\textbf{Overall Pipeline.} 
Given an input image $\mathbf{I}$ and a textual prompt $\mathbf{T}$ (e.g., ``Edema''), we first extract visual features $\mathbf{v} = \Phi_{\text{image}}(\mathbf{I})$ and language embeddings $\mathbf{t} = \Phi_{\text{text}}(\mathbf{T})$. These features are subsequently fused by a cross-attention module \cite{lu2019vilbert} to integrate visual features from chest X-rays with textual embeddings of clinical findings, producing cross-attention maps $\{\mathbf{M}_c\}$, where each $\mathbf{M}_c \in \mathbb{R}^{h \times w}$ corresponds to a particular class label $c$. 

\textbf{Attention Alignment.}
To address the discrepancy between clinicians' attention and the model's attention, the \textit{Attention Aligner} module refines each cross-attention map. We excluded attention loss computation for negative samples because it has been shown that transformers can still focus on specific image regions\cite{abnar2022quantifying} even when no relevant features are present.
Consequently, the refined maps are supervised using two loss terms that are computed only on positive samples—that is, only on the pixels where the ground-truth annotation is non-zero. Let $\Omega^+ = \{ i \in \Omega \mid Y_i \neq 0 \}$ denote the set of positive pixels.

First, for attention alignment and following \cite{luo2024dwarf}, we use a modified dice loss with false positive suppression defined as
\begin{equation}
\mathcal{L}_{\text{AL}} 
= 1-\frac{2 \sum_{i \in \Omega^+} Y_i P_i + \alpha + \varepsilon}
       {\sum_{i \in \Omega^+} (Y_i + P_i) 
        + (w_{\text{FP}} - 1) \sum_{i \in \Omega^+} \text{FP}_i
        + \alpha + \varepsilon}
\end{equation}

where $Y_i$ and $P_i$ denote the Human-annotated ground-truth and predicted attention values at pixel $i$, respectively. The terms $\alpha$ and $\varepsilon$ are smoothing terms, and $w_{\text{FP}}$ is a weighting factor for false positives.

The term $\mathcal{L}_{\text{AL}}$  enforces attention alignment between the provided Human-based attention and the model's attention map, indirectly enforcing the model to learn features yielding similar attention behavior as for the Human expert.







\textbf{Classification Learning via Cross-Entropy Loss.}
This corresponds to the traditional Cross-Entropy loss term used to learn to solve the main task of classification. Let $\mathbf{y}_c \in \{0,1\}$ denote the ground-truth label for finding $c$, and let $\mathbf{z}_c$ be the corresponding logit output from the classification head. Hence, the probability of image $\mathbf{I}$ to be classified as class $c$ is $\mathbf{p}_c = \sigma(\mathbf{z}_c)$, where $\sigma(\cdot)$ denotes the sigmoid function. The cross-entropy loss is computed as $\mathcal{L}_{\mathrm{CE}} 
= - \sum_{c \in \mathcal{N}} \left[
    \mathbf{y}_c \log (\mathbf{p}_c) + (1 - \mathbf{y}_c)\log (1 - \mathbf{p}_c)
\right].$
%
This loss aligns the classification predictions with the ground-truth class labels, improving the model's diagnostic performance.
The final loss is constructed as follows $\mathcal{L}_{\text{total}}
= \mathcal{L}_{\text{CE}}+ \mathcal{L}_{\text{AL}}
$

\begin{figure}[htb]
    \setlength{\belowcaptionskip}{0.1cm}
    \centering
    \includegraphics[width=12.5cm]{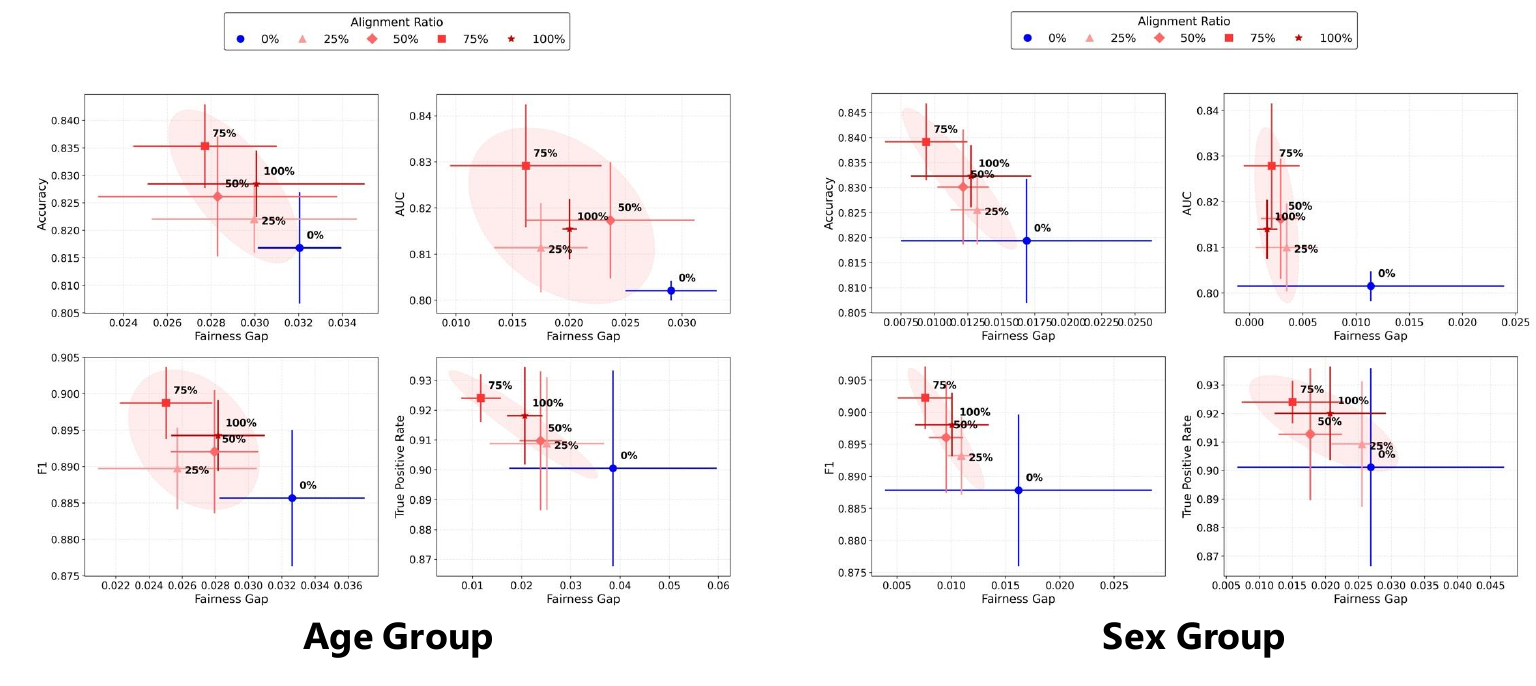}
    \caption{Fairness-performance trade-off for age and sex groups across five levels of Human-AI alignment. Blue points represent non-aligned models ($0\%$), while red-shaded points ($25\%$–$100\%$) indicate increasing alignment. Error bars show variability, and red-shaded ellipses highlight the trends. Fairness improves up to $75\%$ alignment but degrades at $100\%$, suggesting an overconstraining effect. 
    }
    \label{fig:fairness_vs_performance}
\end{figure}

\textbf{Training Details:} For training, we used ViT-B \cite{dosovitskiy2020image} as the visual backbone on an image size of 224 and Med-KEBERT \cite{zhang2023knowledge} as the textual backbone. The model was optimized with AdamW using a learning rate of \( 5 \times 10^{-5} \). The training was conducted on a single H100 96G GPU with a total batch size of 32 for up to 1000 epochs, applying early stopping with patience of 30 epochs. The best-performing model was selected based on the highest validation AUC score. $w_{\text{FP}}$ is set as 2.0. Each experiment was repeated five times, with all reported values averaged over the runs.
 
\section{Results} 
\begin{figure}[htb]
    \centering
    \begin{minipage}[t]{0.47\linewidth}
        \centering
        \includegraphics[width=\linewidth]{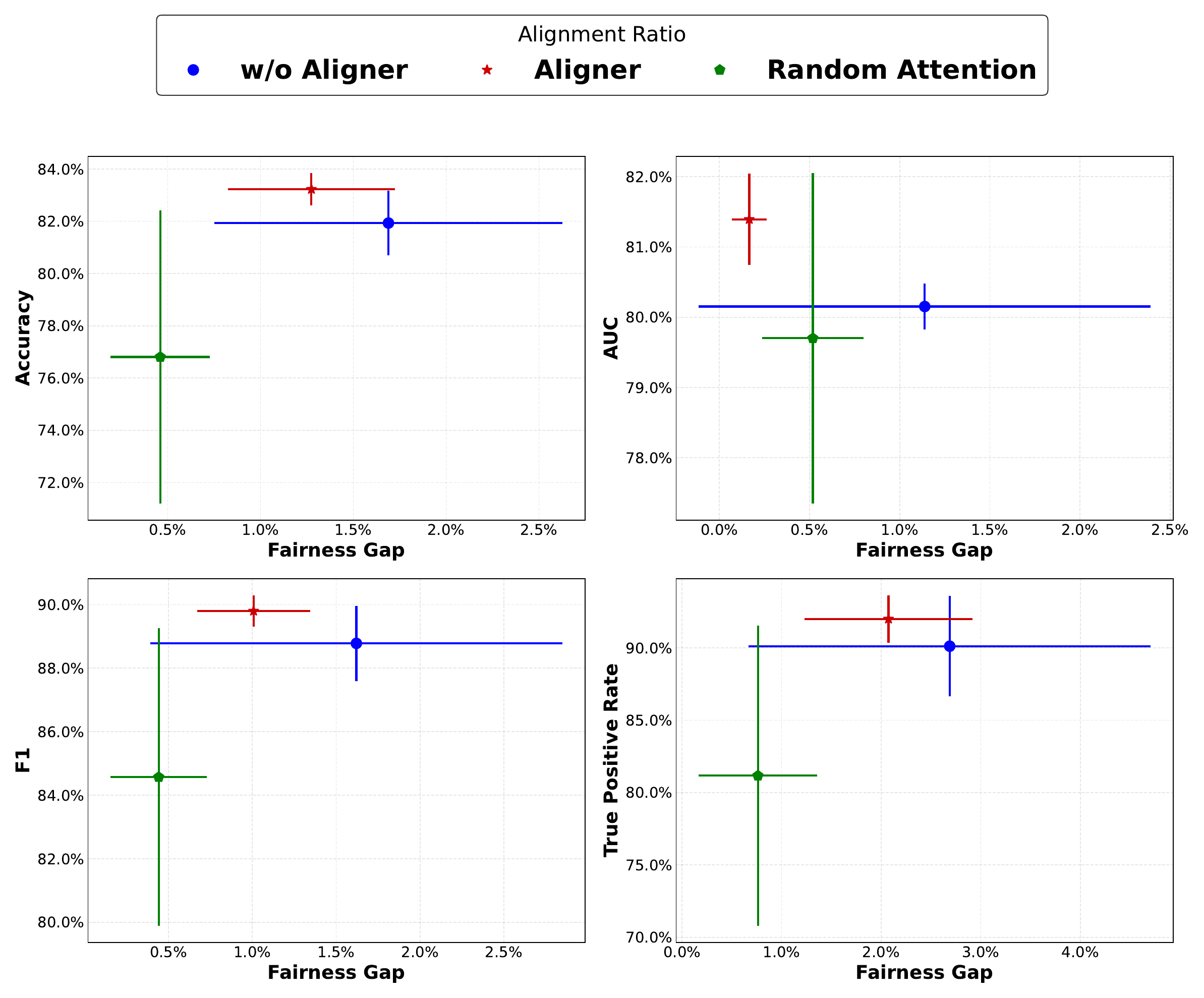}
        \caption{Fairness-performance trade-off results under randomized Human-AI alignment (green) for sex group demographics, compared to a baseline model without alignment (blue), and with Human-AI alignment (red). Error bars represent variability.}
        \label{fig:fake_attention}
    \end{minipage}%
    \hfill 
    \begin{minipage}[t]{0.47\linewidth}
        \centering
        \includegraphics[width=\linewidth]{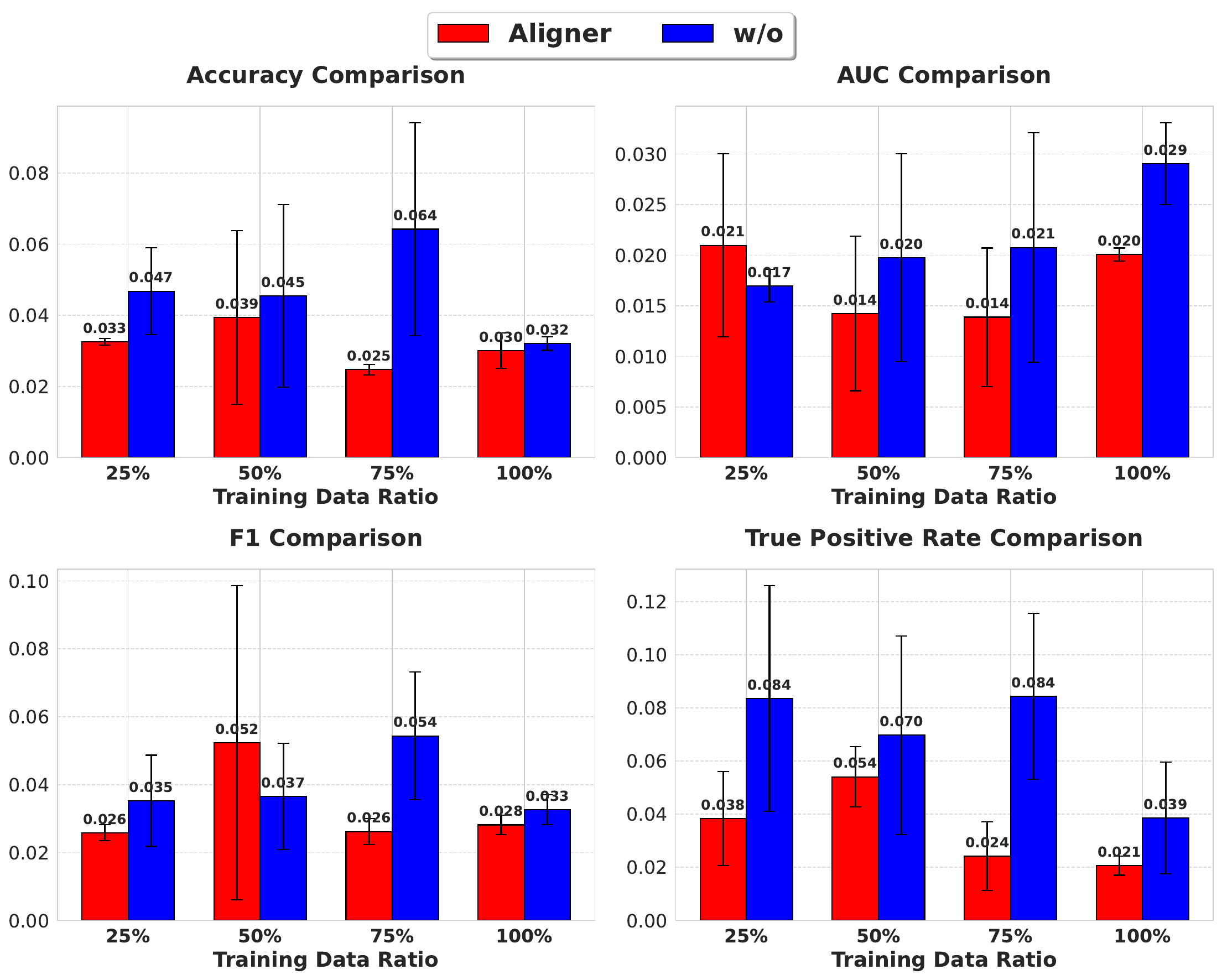}
        \caption{Fairness comparison of Human-AI aligned (red) and non-aligned (blue) models across training data ratios and performance metrics. Lower values indicate better fairness. Alignment reduces fairness gaps, especially in low-data settings}
        \label{fig:low_resource}
    \end{minipage}
\end{figure}

\begin{figure}[htb]
    \setlength{\belowcaptionskip}{0.1cm}
    \centering
    \includegraphics[width=\linewidth]{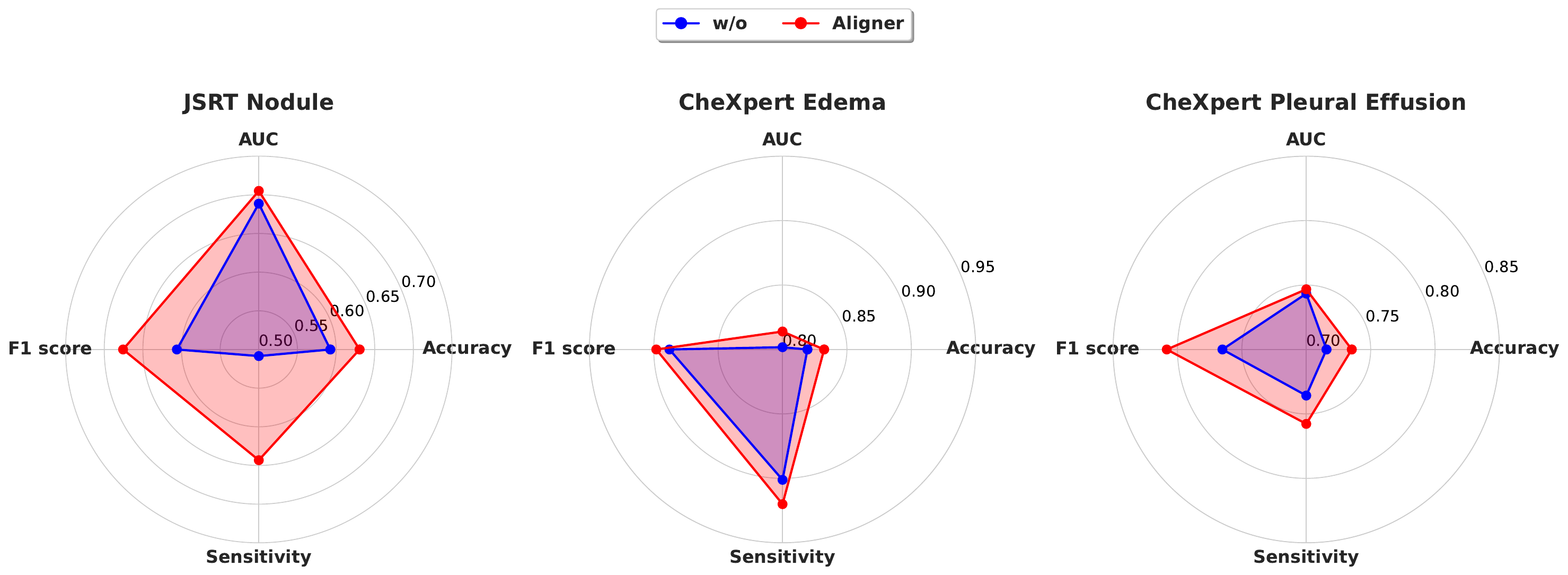}
    \caption{Effect of Human-AI alignment on out-of-domain samples for nodule detection, Edema, and Pleural Effusion. Radar charts compare four performance metrics (outwards is better) with and without alignment. Results show consistent performance improvements across out-of-domain datasets.}
    \label{fig:radar_chart}
\end{figure}

\begin{table}[!hbp]
    \centering
    \setlength{\belowcaptionskip}{0.2cm}
    \caption{Fairness gap comparison between the baseline model (w/o) and the Human-AI aligned model (Aligner) across OOD datasets and two demographic groups. Human-AI alignment reduced fairness gaps in 27 out of 30 comparisons with notable improvements across metrics (lower is better). The Hit Rate indicates the degree of Human-AI alignment (Higher is better). Best metrics in bold.}
    \label{tab:fairness_gap}
    \resizebox{1\columnwidth}{!}{
    \begin{tabular}{cccccccc}
    \toprule
        \textbf{Method} & \textbf{Dataset} & \textbf{Demographic group} & \textbf{Accuracy gap (\%) $\downarrow$} & \textbf{AUC gap$\downarrow$} & \textbf{Sensitivity gap$\downarrow$ } & \textbf{F1 Score gap$\downarrow$} & \textbf{Hit Rate$\uparrow$}  \\
    \midrule
       w/o \cite{zhang2023knowledge} & CheXpert Edema& Age group & 3.20 $\pm$ 0.19 & 2.91 $\pm$ 0.40 & 3.86 $\pm$ 2.11 & 3.26 $\pm$ 0.44& 3.03 $\pm$ 1.50  \\
        Aligner &CheXpert Edema & Age group & 
    \textbf{3.01 $\pm$ 0.57} & \textbf{2.01 $\pm$ 0.07} & \textbf{2.07 $\pm$ 0.42} & \textbf{2.82 $\pm$ 0.33} &  \textbf{14.47 $\pm$ 8.71} \\
    \cmidrule(lr){3-8}
       w/o \cite{zhang2023knowledge} & CheXpert Edema& Gender group & 1.69 $\pm$ 0.94 & 1.14 $\pm$ 1.25 & 2.69 $\pm$ 2.02 &  1.62 $\pm$ 1.23 &  3.03 $\pm$ 1.50 \\
        Aligner & CheXpert Edema& Gender group & \textbf{1.27 $\pm$ 0.45} & \textbf{0.17 $\pm$ 0.10} & \textbf{2.07 $\pm$ 0.84} & \textbf{1.01 $\pm$ 0.34} &  \textbf{14.47 $\pm$ 8.7}  \\
    \midrule
      w/o \cite{zhang2023knowledge} & JSRT Nodule& Age group & \textbf{16.11 $\pm$ 2.42} & \textbf{10.47 $\pm$ 2.16} & 38.28 $\pm$ 9.69 & 33.29 $\pm$ 9.01 & 14.47 $\pm$  8.71 \\
        Aligner  & JSRT Nodule& Age group & 20.09 $\pm$ 4.07 & 11.40 $\pm$ 3.31 & \textbf{28.52 $\pm$ 5.11} & \textbf{25.98 $\pm$ 7.49} & \textbf{22.83 $\pm$  7.96} \\
            \cmidrule(lr){3-8}
       w/o \cite{zhang2023knowledge} &JSRT Nodule & Gender group & \textbf{7.97 $\pm$ 3.30} & 3.50 $\pm$ 2.31 & 12.91 $\pm$ 3.98 & 6.02 $\pm$ 3.52 & 14.47 $\pm$  8.71 \\
        Aligner &JSRT Nodule & Gender group & 8.25 $\pm$ 4.31 & \textbf{1.07 $\pm$ 1.15} & \textbf{9.59 $\pm$ 9.20} & \textbf{7.68 $\pm$ 4.70} & \textbf{22.83 $\pm$  7.96} \\
    \midrule
       w/o \cite{zhang2023knowledge} & CheXpert Pleural Effusion& Age group & 9.50 $\pm$ 1.82 & 3.20 $\pm$ 0.48 & 17.42 $\pm$ 3.29 & 12.39 $\pm$ 3.52 & 10.82  $\pm$ 4.56 \\
        Aligner   &CheXpert Pleural Effusion& Age group & \textbf{6.14 $\pm$ 0.38} & \textbf{3.84 $\pm$ 0.16} & \textbf{11.28 $\pm$ 1.02} & \textbf{7.51 $\pm$ 0.73} &  \textbf{24.24 $\pm$ 13.33} \\
            \cmidrule(lr){3-8}
       w/o \cite{zhang2023knowledge} & CheXpert Pleural Effusion & Gender group & 1.50 $\pm$ 0.99 & \textbf{0.21 $\pm$ 0.18} & 2.09 $\pm$ 1.24 & 1.30 $\pm$ 0.98 & 10.82  $\pm$ 4.56 \\
        Aligner &CheXpert Pleural Effusion & Gender group &\textbf{0.83 $\pm$ 0.93} & 0.23 $\pm$ 0.20 & \textbf{1.51 $\pm$ 1.19} & \textbf{0.64 $\pm$ 0.68} &  \textbf{24.24 $\pm$ 13.33} \\
    \bottomrule
    \end{tabular}}
\end{table}

\textbf{Human-AI alignment improves fairness gap among different demographic groups:}
Table~\ref{tab:fairness_gap} shows the main results of our study assessing the impact of Human-AI alignment on fairness among demographic groups and performance across three datasets (CheXpert Edema, JSRT Nodule, and CheXpert Pleural Effusion). 
The results demonstrate that, compared to the baseline model without alignment (i.e., labeled as \textit{w/o} in Table~\ref{tab:fairness_gap}), Human-AI alignment improves the fairness gap for sex and age groups across different performance metrics and datasets, with improvements in 27 out of 30 comparisons (i.e., 5 metrics $\times$ 3 datasets $\times$ 2 demographic groups, Table~\ref{tab:fairness_gap}). Similar trends are observed in Fig.~\ref{fig:fairness_vs_performance}, showing that Human-AI alignment improves fairness metrics. However, exacerbating the alignment can also lead to diminished gains or even unintended trade-offs in fairness and performance. This finding aligns with the recent observations reported in \cite{gorbatovski2024learn}.  

\textbf{Human-AI alignment improves performance in out-of-domain samples:} Figure~\ref{fig:radar_chart} shows the effect of Human-AI alignment on out-of-domain samples for nodule and mass detection, Edema, and Pleural Effusion (Table \ref{tab:TrainOODDataset}). Each radar chart shows the four performance metrics (higher the better), with and without human-AI alignment. Results show considerable performance improvements suggesting that Human-AI alignment promoted not only fairness improvements but also performance improvement on out-of-domain datasets, reflecting an important property for real-world clinical scenarios.

\textbf{Human-AI alignment ensures stable fairness improvements in low-data scenarios:} Figure~\ref{fig:low_resource} shows that Human-AI alignment improves fairness across all training data ratios, with the most significant impact in low-data scenarios (25\%–50\%), where it helps mitigate disparities more effectively.

\textbf{Randomized Human-AI alignment reduces performance and fairness gap:} Figure~\ref{fig:fake_attention} presents the fairness-performance trade-off when Human-AI guidance is randomized (green points, the generation of random attention is illustrated at Fig\ref{fig:studydesign}). As expected, randomization degrades performance but also reduces fairness gaps, suggesting a decorrelation effect on demographic attributes. This trade-off aligns with fairness-aware modeling literature, where reducing bias can sometimes come at the cost of lower performance.

\section{Conclusion}
Our study provides the first systematic exploration of the interplay between Human-AI alignment and fairness in medical image classification. Our results demonstrate that \emph{Human-AI alignment consistently reduces fairness gaps} across sex and age groups, with improvements observed across datasets, tasks, and performance metrics, reinforcing the robustness of these findings.
Beyond fairness benefits, we found that \emph{Human-AI alignment enhances out-of-domain performance}, an essential property for real-world clinical deployment. These gains suggest that aligning model representations with human knowledge not only reduces bias but also strengthens performance when applied to unseen data, challenging the notion that fairness-improving interventions necessarily degrade accuracy.
However, our findings also highlight the need for careful design and calibration of alignment strategies. While alignment generally improves both fairness and performance, \emph{excessive alignment can lead to diminished gains or even unintended trade-offs}. Our randomized alignment ablation study further revealed that \emph{misguided alignment degrades performance while also reducing fairness gaps}, suggesting a decorrelation effect between model predictions and demographic attributes. These results emphasize that the effectiveness of fairness interventions depends on how they are applied, underscoring the importance of balancing alignment for fairness and model utility. Overall, these findings highlight Human-AI alignment as a promising avenue for developing fair, robust, and generalizable AI models in medical imaging.

\FloatBarrier
\bibliographystyle{splncs04}
\bibliography{egbib}
\end{document}